\documentclass[sigconf]{acmart}
\usepackage{tabularx}
\usepackage{makecell} 
\usepackage{bchart}
\usepackage{pgfplots}
\usepackage{xcolor}
\definecolor{mintgreen}{RGB}{135,232,133}
\definecolor{darkblue}{RGB}{255,192,118}
\definecolor{orange}{RGB}{250,128,128}
\AtBeginDocument{%
  }

\setcopyright{acmcopyright}
\copyrightyear{2023}
\acmYear{2023}
\acmDOI{XXXXXXX.XXXXXXX}

\acmConference[KDD-UC '23]{KDD Undergraduate Consortium}{August 06--10,
  2023}{Long Beach, CA}
\acmPrice{15.00}
\acmISBN{978-1-4503-XXXX-X/23/08}

\begin{document}

\title{ViT2EEG: Leveraging Hybrid Pretrained Vision Transformers for EEG Data}

\author{Ruiqi Yang}
\authornote{Both authors contributed equally to this research.}
\email{ruiqiyang@ucsb.edu}
\affiliation{%
  \institution{University of California Santa Barbara}
  \city{Santa Barbara}
  \state{CA}
  \country{USA}
  \postcode{93117}
}

\author{Eric Modesitt}
\authornotemark[1]
\email{ericjm4@illinois.edu}
\affiliation{%
  \institution{University of Illinois at Urbana Champaign}
  \city{Urbana-Champaign}
  \state{Illinois}
  \country{USA}
  \postcode{61801}
}

\renewcommand{\shortauthors}{Ruiqi Yang and Eric Modesitt}

\begin{abstract}
In this study, we demonstrate the application of a hybrid Vision Transformer (ViT) model, pretrained on ImageNet, on an electroencephalogram (EEG) regression task. Despite being originally trained for image classification tasks, when fine-tuned on EEG data, this model shows a notable increase in performance compared to other models, including an identical architecture ViT trained without the ImageNet weights. This discovery challenges the traditional understanding of model generalization, suggesting that Transformer models pretrained on seemingly unrelated image data can provide valuable priors for EEG regression tasks with an appropriate fine-tuning pipeline.

The success of this approach suggests that the features extracted by ViT models in the context of visual tasks can be readily transformed for the purpose of EEG predictive modeling.
We recommend utilizing this methodology not only in neuroscience and related fields, but generally for any task where data collection is limited by practical, financial, or ethical constraints.
Our results illuminate the potential of pretrained models on tasks that are clearly distinct from their original purpose.
\end{abstract}


\maketitle

\section{Introduction}
With its rich, multidimensional structure, electroencephalogram (EEG) data encapsulates a wealth of information about brain activity, offering unique insights into a variety of neurological phenomena \cite{teplan2002fundamentals}. However, the intricacies that define EEG data introduce substantial hurdles towards devising efficient, yet effective predictive models, especially considering the incredibly costly process of data collection \cite{craik2019deep}. Despite the extensive usage of machine learning regression models for predicting EEG data, these models often fall short of adequately capturing and decoding the complex structures within such data. As an example, the EEGEyeNet dataset \cite{kastrati2021EEGEyeNet} provides a baseline for predicting the position of a subject's gaze on a screen given EEG data. Their results clearly demonstrate the inability of machine learning models to accurately interpret the given data. We also explore articles that employ analogous time-series data \cite{dou2022time,lu2023machine,murungi2023trends,qu2020multi,qu2020identifying,qu2020using,tang2023active,yi2022attention}. 

In recent years, there has been a significant surge in the invention and application of deep learning models to unravel this complexity and extract valuable insights from EEG data. In particular, Convolutional Neural Networks (CNNs) and \cite{lecun1998gradient} the Self-Attention operator \cite{vaswani2017attention} have demonstrated promising results in managing the complexity of EEG data \cite{roy2019deep, altaheri2021deep}, with several architectures being derived from their successful counterparts in the field of Computer Vision \cite{khan2022transformers,wang2018non, cao2019gcnet}. These deep learning techniques excel at capturing nonlinear patterns and relationships within data, making them particularly suitable for modeling complex EEG signals.

Our study takes the paradigm of converting successful computer vision models to an EEG counterpart a step further and proposes the use of a hybrid Vision Transformer (ViT) \cite{dosovitskiy2020image} pretrained on the ImageNet \cite{deng2009imagenet} dataset to take on an EEG regression task. This approach takes a theoretical leap from previous methods, repurposing a model's architecture and weights initially designed for vision tasks to grapple with the high dimensionality and inherent complexity of EEG data. This research is predicated on the observation that EEG data, like image data, possesses rich, high-dimensional structures that can be effectively modeled using Transformer models \cite{vaswani2017attention}.

Our empirical findings indicate that utilizing a pretrained ViT model significantly outperforms traditional regression models. In addition, we demonstrate that the success of the model comes from not just the model architecture, but also the impact of pretraining the ViT on the model's initial weights. This success underlines the strength of pretraining the model across disciplines, demonstrating how models trained on abundant, easily obtained image data can be effectively repurposed for EEG data analysis, an area where data collection often involves considerable challenges due to practical, financial, and ethical constraints \cite{roy2019deep}.

\section{Related Work}

\subsection{Deep Learning for EEG Tasks}

EEG regression and classification tasks, fundamental to neuroscience research and applications like Brain-Computer Interfaces (BCIs), have traditionally relied on Convolutional Neural Networks (CNNs) \cite{dai2020HSCNN,zhou2018epileptic}. While CNNs have proven their use for extracting localized features from time-series EEG data, they struggle with capturing the long-term dependencies, subject-independent and session-independent patterns inherent in EEG data \cite{altaheri2021deep}. This limitation motivates the exploration of other model architectures that can replace or support CNNs.

Recently, hybrid methods complementing CNNs with Self-Attention modules \cite{bello2019attention} have shown promising improvements in performance by capturing both local patterns and long-term dependencies in EEG signals \cite{tao2020EEG, xiao20224D}. In addition, recent studies in Computer Vision have demonstrated the superiority of  Transformer-focused models over predictive models focused on convolutional layers \cite{touvron2021training,wang2021pyramid}. Following this trend, our study aims to test the standalone capabilities of Transformers (specifically Vision Transformers), expecting the ViT's ability to capture global dependencies in the data to deliver superior performance in the complex, high-dimensional EEG data present in eye-tracking tasks.

\subsection{EEG Eye-Tracking with EEGEyeNet}

Using EEG data for applications in eye-tracking presents unique challenges and applications, ranging from assisting with cognitive studies to developing assistive technologies. The recent introduction of the EEGEyeNet dataset \cite{kastrati2021EEGEyeNet}, containing 47 hours of high-density 128-channel EEG data synchronized with eye-tracking recordings from 356 healthy adults, offers a valuable resource for studying such complex relationships. This dataset not only aids in studying attention and reaction time, but it also provides a preprocessing pipeline and benchmarks for gaze estimation, facilitating reproducible research every step of the way.

\subsection{Vision Transformers (ViTs)}

Introduced by \cite{dosovitskiy2020image}, Vision Transformers (ViTs) have demonstrated impressive results in image classification tasks. Unlike CNNs, which use spatial convolution operations for feature extraction, ViTs divide the input image into a grid of patches and leverage transformer architecture to process these patches as a sequence. This design allows ViTs to capture global dependencies in the input data without convolution operations,  making them more suited for the globally-correlated nature of EEG data.

Research in similar domains indicates the applicability of ViTs. For example, studies have demonstrated ViT's effectiveness in tasks involving audio \cite{gong2021ast} and video processing \cite{zeng2020learning, arnab2021vivit}, hinting at its potential for spatial-temporal analysis of EEG data. Despite its potential, the usage of ViTs in EEG analysis is largely unexplored, prompting the need for research like the current study.

\subsection{Pretrained ViTs}
The innate capacity of the self-attention mechanism in Vision Transformer (ViT) models to possess a global receptive field facilitates simultaneous contemplation of all input fragments during decision-making procedures. This particular ability is critical in capturing intricate and distant dependencies, a characteristic frequently encountered in Electroencephalogram (EEG) data. Moreover, the transferability of pretrained ViT models \cite{yuan2021tokens,han2021transformer,liu2021swin}, which are originally trained on vast datasets, permits their fine-tuning for subsequent tasks. This property is particularly advantageous for EEG-related endeavors, which often grapple with a dearth of labeled data.

Nonetheless, despite the demonstrable success and potential of pretrained ViT models, it is imperative to recognize their limitations when deployed directly to realms that are substantially distinct from their original training context. One such example is the inherently high-dimensional and noisy nature of EEG data. Hence, the present investigation critically assesses the efficacy of employing a pretrained ViT model for an EEG regression task. This exploration could consequently lay a foundation for future scientific inquiries in this domain.

\begin{figure}[t]
  \centering
  \includegraphics[width=\columnwidth]{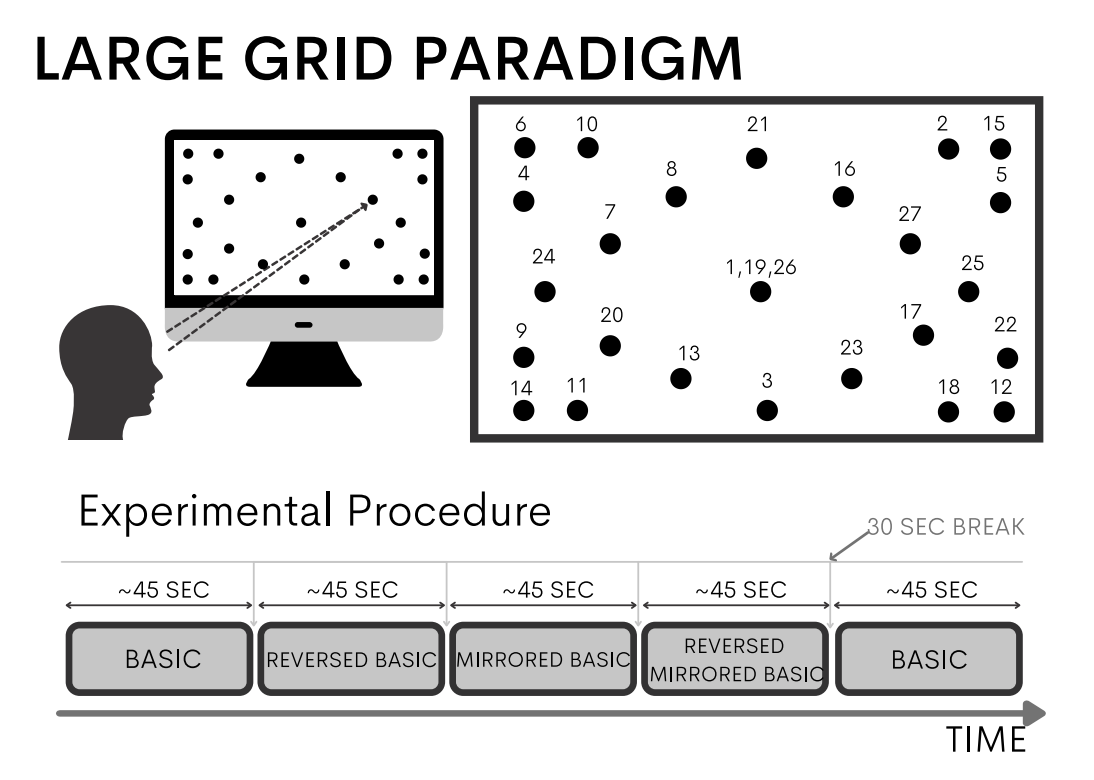}
  \caption{EEGEyeNet Large Grid Paradigm \cite{kastrati2021EEGEyeNet}. Participants are asked to fixate on particular dots in a given period.}
  \label{fig:largegrid}
\end{figure}

\section{Methods}

\begin{table}[b]
\centering
\begin{tabular}{lcccccccc}
\hline
\multicolumn{4}{c}{\textbf{Participants} } \\
Total & Train & Validation & Test  \\ 
\hline
27 & 19 & 4 & 4 \\
\hline
\multicolumn{4}{c}{\textbf{Samples} } \\
 Total & Train & Validation & Test  \\ 
 \hline
  21464 & 14706 & 3277 & 3481 \\

\hline
\end{tabular}
\caption{Distribution of participants and samples in the Abs. Position dataset.}
\label{tab:dataset_distribution}
\end{table}

\subsection{Structuring Images from EEG Time Series Data}

\begin{figure*}[t]
  \centering
  \includegraphics[width=\textwidth]{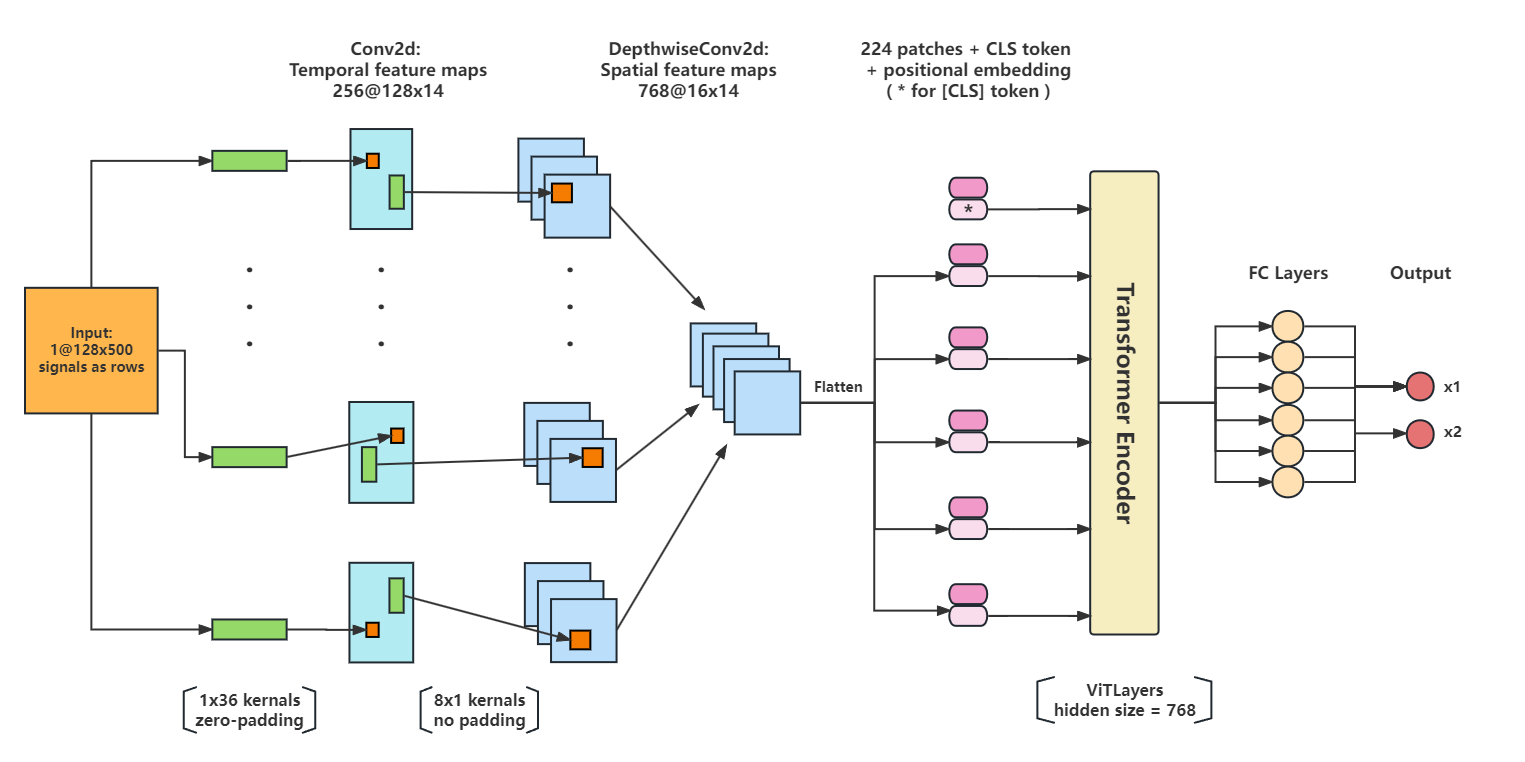}
  \caption{Proposed EEGViT, a hybrid ViT (Vision Transformer) architecture designed specifically for EEG raw signal as input.  A two-step convolution operation is applied to generate patch embeddings. Then we add positional embeddings and pass the resulting sequence into ViT layers. The illustration of positional embedding and ViT layer is based on \cite{dosovitskiy2020image}.}
  \label{fig:eegvit}
\end{figure*}

The data utilized in this research comes directly from the large grid paradigm of EEGEyeNet (see Fig. \ref{fig:largegrid}). Participants of this paradigm are instructed to fixate on a sequence of 25 dots appearing at distinct screen positions, for 1.5 to 1.8 seconds each. We primarily used data from EEGEyeNet's Absolute Position Task, which concentrates on pinpointing the precise XY-coordinates of the subject's gaze on the screen \cite{kastrati2021EEGEyeNet}.

The dataset consists of raw EEG signals collected by a 128-channel Geodesic Hydrocel system at sampling rate of 500Hz, and it undergoes a preprocessing process recommended by \cite{hollenstein2018zuco}. Initially, defective electrodes that could potentially introduce noise were identified and handled. Following this, the data underwent frequency filtering, with a 40 Hz high-pass filter and a 0.5 Hz low-pass filter applied, which allowed us to retain useful frequencies and eliminate potential noise. Eventually, the resulted signals were segmented into 1-second trials including 500 time samples for each of 128 channels, and we stacked channels row by row to form a $128\times500$ matrix as one input trial.

The processed data was then divided into training, validation, and test sets in a 70\%, 15\%, 15\% distribution. The validation and test data sets were made up of distinct subjects, as depicted in Table \ref{tab:dataset_distribution}. This approach allowed us to assess the model's ability to generalize and adapt to new data, providing a more robust and accurate evaluation of its potential real-world application.

\subsection{Model Architecture}
The model presented in this paper, named EEGViT, is a hybrid Vision Transformer architecture that integrates a two-step convolution operation during the patch embedding process (Fig. \ref{fig:eegvit}).

\subsubsection{Two-Step Convolution Block}

\begin{table}[b]
\hspace*{-0.2cm}
\centering
\begin{tabular}{lccc}
\hline
 Layers & Kernel & Channels & Options  \\ 
\hline
Conv2D & $1\times36$ & 256 & padding = (0,2)\\
       & & & stride = (1,36)\\
\hline
BatchNorm2D & & 256\\
\hline
DepthwiseConv2D & $8\times1$ & 768 & stride = (8,1)\\
\hline
\end{tabular}
\caption{Details of Two-Step Convolution Block in EEGViT, implemented with PyTorch library \cite{paszke2019pytorch}}
\label{tab:conv_layer}
\end{table}

We adopt the two-step convolution block utilized in EEG analysis from the early layers of EEGNet \cite{lawhern2018EEGNet}. This block consists of two convolutional layers, with one layer filtering the temporal dimension and the other layer filtering the channel (spatial) dimension.

The first layer employs a $1 \times T$ kernel and scans across the entire input, capturing temporal events that occur over the same channels. The kernels in this temporal convolutional layer can be regarded as various band-pass filters applied to the raw signals. The resulting output feature maps represent the frequency bands of interest for the EEG. Batch normalization is applied on the output \cite{ioffe2015batch}.

Following this, a depthwise convolutional layer \cite{howard2017mobilenets} containing a $C \times 1$ kernel is used. This layer scans over multiple channels, given the same point in time, filtering the inputs separately. This depthwise convolution also acts as a frequency-specific spatial filter when applied to EEG feature extraction in this context.

The overall operation can be viewed as splitting the input images into $C \times T$ patches, followed by a row-by-row linear projection, with the resulting column vector transformed into a scalar feature. The decision to separate the projection for each dimension and incorporate the depthwise mechanism proves advantageous.

To align with the ViT-base model's encoder layers, which are pretrained on images divided into 192 patches \cite{dosovitskiy2020image}, we carefully select hyperparameters for an output of 224 patch embedding vectors. These patches, each representing the activity of eight adjacent channels over 36 timestamps (72ms at a 500Hz sampling rate), are set to be non-overlapping, based on experimental findings. Hyperparameters are further tuned for optimal feature preservation and computational efficiency (Table \ref{tab:conv_layer}).

Lastly, to integrate our patch sequence into the pretrained ViT layers, which use 192 patches and corresponding pretrained positional embeddings, we reinitialize a sequence of positional embeddings to match our patch sequence length."

\subsubsection{Transformer Block}

\begin{table}[b]
\hspace*{-0.2cm}
\centering
\begin{tabular}{lcccc}
\hline
 & Modules & Hidden Size & MLP Size & Heads \\ 
\hline
ViT-Base & 12 & 768 & 3072 & 12  \\
\hline
\end{tabular}
\caption{Details of ViT-Base model \cite{dosovitskiy2020image} adopted for the transformer block in EEGViT}
\label{tab:transformer_block}
\end{table}

\begin{figure}[t]
  \centering
  \hspace*{-1cm}
  \includegraphics[width=0.38\columnwidth]{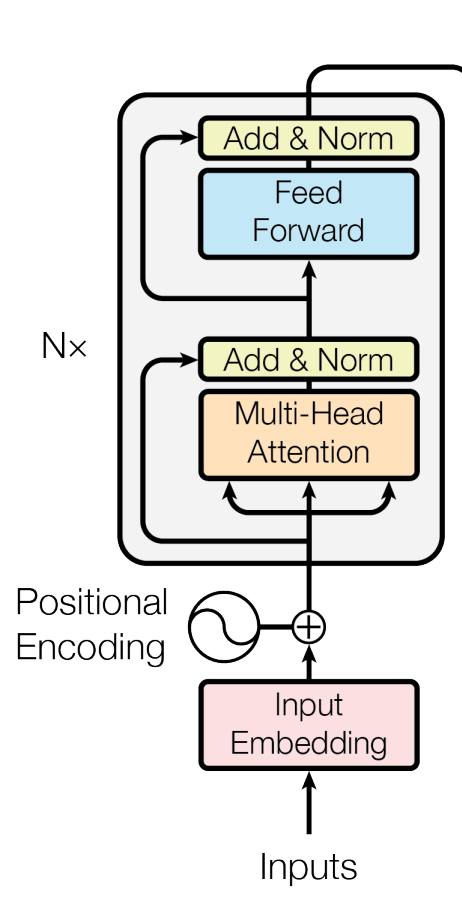}
  \caption{Transformer Encoder block by \cite{vaswani2017attention} We follow this paradigm to implement the Transformer block in EEGViT.}
  \label{fig:vit_encoder}
\end{figure}

We adopt the transformer encoder paradigm (see Fig. \ref{fig:vit_encoder}) and utilize the ViT-Base model architecture in \cite{dosovitskiy2020image} to implement the transformer block in our model. This block consists of 12 modules, each comprising a Multi-Head Attention layer and a Multi-Layer Perceptron,  wrapped by residual connection and layer normalization. Detailed hyperparameters for these layers can be found in Table \ref{tab:transformer_block}.

For the pretrained version of the transformer block, we load the weights of the encoder layers from the ImageNet-pretrained ViT-Base model, consisting of approximately 86M parameters, and readily available on HuggingFace \cite{wolf2020transformers}.

The input to the transformer block is a sequence of 1D patch embeddings obtained by flattening the output feature maps of the preceding Two-Step Convolution block. We incorporate positional embeddings into this sequence. Additionally, inspired by ViT, we adopt BERT's approach of appending a CLS token to the sequence. This CLS token aids in aggregating information across the sequence \cite{devlin2018bert}. The hidden state of the CLS token ultimately serves as the representation of the entire sequence of patch embeddings and is subsequently transformed into the target space, which corresponds to 2D coordinates on the grid, using fully connected layers.

\subsection{Training}

Training machine learning models effectively requires a meticulous and strategic approach to setting various parameters and configurations. The goal is to optimize the model's learning process, enabling it to understand complex patterns in the data and deliver reliable predictions on new, unseen data. The EEGViT model was trained with such attention to detail.

For our study, we adopted the Mean Squared Error (MSE) loss function as the primary measure to guide our model's learning. MSE loss function is widely used in regression problems due to its mathematical simplicity and effectiveness. In our specific context of EEG signal analysis, it allows the model to minimize the prediction error of continuous output values, essentially improving the model's ability to predict the exact EEG signal values. Although we employed MSE for training, we decided to use Root Mean Square Error (RMSE) to measure our results. 

It's essential to mention that using MSE as the loss function and subsequently using RMSE as the performance metric is a commonly adopted approach in various scientific studies, particularly those involving EEG and eye-tracking data \cite{kastrati2021EEGEyeNet}. The use of these metrics is not only restricted to our work but has also been used extensively in similar studies.

This common usage is mainly due to the advantage that these metrics offer in evaluating the model's ability to accurately predict continuous outcomes while minimizing the overall deviation. Moreover, since these metrics are squared, they place a higher penalty on larger errors, thereby pushing the model to improve its overall accuracy. This attribute is extremely beneficial in sensitive analyses like EEG signal interpretation, where the aim is to minimize prediction discrepancies as much as possible.

Furthermore, the use of RMSE as a performance metric provides us a direct comparison with other research findings in the field of EEG eye-tracking. This allows for a standardized evaluation across various studies, enhancing the generalizability and transferability of our findings within the field. Given that other studies have successfully employed the same method, it strengthens our confidence in the adopted approach and provides more credibility to our results.

In order to compare different strategies and evaluate the effectiveness of our proposed approach, we trained two different model architectures. The first was the unmodified, non-pretrained ViT-Base model. This model, serving as a benchmark, was a simple architecture comprised of a single convolutional layer that used 8x36 kernels (i.e. patches) to generate patch embeddings. Alongside this plain version, we used a pretrained version of the ViT-Base model trained on the ImageNet dataset \cite{deng2009imagenet}, using the weights from \cite{wolf2020transformers}. This pre-training approach aimed to utilize any cross-image features the ViT learned from the vast ImageNet dataset that may be applicable to EEG data. 

Secondly, we modified the ViT-Base model into the proposed EEGViT model by adding the Two-Step Convolution Block. We trained two versions of the EEGViT model, one using randomly initlized weights and the other relying on the weights of ViT-Base obtained on the ImageNet dataset. This in effect allows for our EEGViT model to be comparable to both versions of the ViT-Base model.

Each of these models underwent training for 15 epochs with one RTX 4090 GPU. Each epoch takes about one minute to iterate through all the batches of 64 samples. For the models without pre-training, we kept the learning rate at a steady value of $10^{-4}$ throughout the training process. For the pretrained model, we dropped the initial learning rate of 1e-4 by a factor of 0.9 every 6 epochs. This ensured that the models were exposed to a sufficient level of detail in the data to capture intricate patterns without overwhelming the finetuning process with too much noise.

\section{Baseline Methods}

\subsection{Naive Guessing}
In an attempt to compare our performance against a general naive, untrained model, we adopt the Naive Guessing baseline from EEGEyeNet, where the mean position from the training set is used as the constant prediction for every test data point. According to the original EEGEyeNet paper, this yields a RMSE distance of 123.3 mm, a standard we aim to improve upon.

\subsection{Machine Learning Methods}

Machine Learning methods tried in the original EEGEyeNet study did not achieve significant improvements over the naive baseline when estimating absolute position \cite{kastrati2021EEGEyeNet}. In this study, we re-evaluated these methods in the following sections. Details regarding these models can be found in Appendix.

\subsection{Deep Learning Methods}

In this study, we focus on a comparison of the proposed EEGViT with state-of-the-art CNN-based architectures, as they are widely recognized for their effectiveness in handling EEG data and have demonstrated state of the art performance on the EEGEyeNet benchmark.

\textbf{Convolutional Neural Network (CNN)}
We implemented a standard one-dimensional CNN, applying max pooling in line with EEGEyeNet.

\textbf{PyramidalCNN}
The PyramidalCNN model leverages varying time granularity in the CNN architecture to capture temporal features at multiple scales \cite{johnson2017deep}.

\textbf{EEGNet}
EEGNet, an EEG-specific CNN architecture that employs depthwise and separable convolutions, was designed for versatility across Brain-Computer Interface (BCI) paradigms \cite{lawhern2018EEGNet}.

\textbf{InceptionTime}
InceptionTime, a scalable Time Series Classification (TSC) model, employs an ensemble of deep CNNs engineered for TSC tasks \cite{fawaz2020inceptiontime}.

\textbf{Xception}
Xception, an efficient CNN architecture employing depthwise separable convolutions, was designed to capture spatial correlations while enabling faster training \cite{chollet2017xception}.

\begin{table}[t]
\centering
\begin{tabular}{|l|c|}
\hline
\textbf{Model} & \textbf{Absolute Position RMSE (mm)} \\ \hline
Naive Guessing & 123.3 ± 0.0 \\ \hline \hline
KNN & 119.7 ± 0 \\ \hline
RBF SVR & 123 ± 0 \\ \hline
Linear Regression & 118.3 ± 0 \\ \hline
Ridge Regression & 118.2 ± 0 \\ \hline
Lasso Regression & 118 ± 0 \\ \hline
Elastic Net & 118.1 ± 0 \\ \hline \hline
Random Forest & 116.7 ± 0.1 \\ \hline
Gradient Boost & 117 ± 0.1 \\ \hline
AdaBoost & 119.4 ± 0.1 \\ \hline
XGBoost & 118 ± 0 \\ \hline \hline
CNN & 70.4 ± 1.1 \\ \hline
PyramidalCNN & 73.9 ± 1.9 \\ \hline
EEGNet & 81.3 ± 1.0 \\ \hline
InceptionTime & 70.7 ± 0.8 \\ \hline
Xception & 78.7 ± 1.6 \\  \hline \hline
ViT-Base & 61.5 ± 0.6 \\ \hline
ViT-Base Pre-trained & 58.1 ± 0.6 \\ \hline \hline
EEGViT & 61.7 ± 0.6 \\ \hline
\textbf{EEGViT Pre-trained} & \textbf{55.4 ± 0.2} \\ \hline

\end{tabular}

\caption{Comparison of Root Mean Squared Error (RMSE) loss in millimeters for different models on the Absolute Position Task. Original error is in pixels, and we convert it into millimeters by 2 pixels/mm for better interpretation. Lower RMSE values indicate better performance as they represent closer estimations to the actual values. The values represent the mean and standard deviation of 5 runs.}
\label{tab:model_comparison}

\end{table}

\section{Results}

In this study, we set out to evaluate and compare the performance of an array of models for an EEG regression task. Our approach included an assessment of traditional Machine Learning models, Deep Neural Networks, as well as variants of ViT models, both with and without pretraining and the Two-Step convolution block. We gauged the models' performance using the root mean square error (RMSE) on a dedicated test set. Our comprehensive findings, which also encompass the naive baseline, are outlined in Table \ref{tab:model_comparison}.

\subsection{Baseline Regression Models}

The outcomes from these methods establish a baseline that outperforms the naive baseline considerably. However, it is evident that there is ample scope for enhancement. The best-performing model of this group, the Convolutional Neural Network (CNN), could only achieve an average RMSE of 70.4 mm, indicating potential room for improvement.

\subsection{Transformer Models}

Both the basic versions of ViT-Base and our EEGViT demonstrated a similar level of performance, registering a significant 12\% improvement over the preceding state-of-the-art models. This marked improvement underlines the utility of the Attention mechanism in capturing the temporal intricacies in EEG data.

\subsection{Pretrained Transformer Models}

\begin{figure}[t]
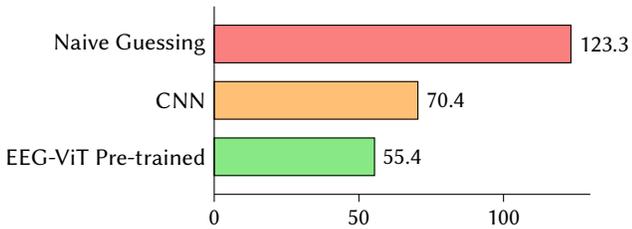

\hskip -0.3cm
\begin{bchart}[max=130,step=50,width=5cm]
    \bcbar[color=orange,label=Naive Guessing]{123.3}
    \smallskip
    \bcbar[color=darkblue,label=CNN]{70.4}
    \smallskip
    \bcbar[color=mintgreen,label=EEG-ViT Pre-trained]{55.4}
    
\end{bchart}
    \caption{Comparison of Root Mean Squared Error (RMSE) loss in millimeters for random guessing, previous state of the art (CNN) and our novel method (EEG-ViT Pre-trained). Lower is better.}
    \label{tab:oursvssota}
\end{figure}

The pretrained versions of both ViT-Base and EEGViT exceeded the performance of the previous models by significantly enhancing accuracy. More specifically, EEGViT demonstrated an exceptional performance boost of 10\% compared to the non-pretrained ViT, and an impressive advancement of about 21\% over the preceding state-of-the-art model, as shown in Figure \ref{tab:oursvssota}.

Interestingly, even though ViT-Base underwent an identical pretraining process as EEGViT, it did not showcase the same level of prowess. ViT-Base's loss on the test set was roughly 5\% higher than EEGViT's, which deviates from their comparative performance when trained from scratch. This disparity can likely be attributed to the positive impact of pretraining.

When the model can draw from a substantial knowledge base, the advantages of enhanced representations become more noticeable. Consequently, the Two-Step convolution representations in our model likely lead to better leverage within the Attention mechanism and Transformer architecture. However, this difference in performance highlights the need for further research into optimizing the training process and leveraging domain-specific knowledge in EEG signal processing.

\section{Discussion}
Our EEGViT model significantly outperforms conventional regression models, Convolutional Neural Networks (CNNs), and the baseline ViT-Base, establishing the utility of transfer learning and the enhancement potential of image data for EEG analysis.

Despite incorporating convolutional layers, our EEGViT model's superior performance is primarily attributable to the Transformer block. This is highlighted when compared with baseline models that do not use a Transformer block, such as EEGNet and Xception, which also employ depthwise convolutions in their early layers, yet underperform relative to EEGViT.

Our findings accentuate the significance of reduced inductive bias in transitioning from Convolutional Neural Networks (CNNs) to Transformer-based architectures for processing extensive datasets. This echoes recent studies within the Computer Vision field, suggesting that the inherent strong inductive bias of CNNs may impede their efficiency, particularly with large-scale data, thereby making Transformers a preferable alternative \cite{d2021convit, dosovitskiy2020image}. Our investigation mirrors this, demonstrating that an unmodified Vision Transformer (ViT) exceeds CNN performance when trained exclusively on the EEGEyeNet dataset, a trend that is amplified with pretraining. Consequently, future research might explore strategies to mitigate the intrinsic inductive bias of prediction models, especially in data-rich environments.

However, there is a compelling need to make these models more interpretable. Future work should consider transforming our model into an interpretable version. For instance, visualization techniques could be incorporated to understand the activation patterns and correlations in the data.

Our EEGViT model holds promise for a wide range of EEG datasets. The EEGEyeNet dataset, upon which this study is based, is an extensive compilation of synchronized EEG and eye-tracking data. Given the very nature of the input data it learns, we strongly believe that the methodology presented in this paper can be applied to other, similar EEG datasets. 

The practical value of our findings lies in the model's ability to supplement scarce EEG data with readily available image data, thereby reducing ethical and financial barriers to EEG data collection and analysis. This approach could streamline research in the field, facilitating robust, scalable EEG studies.

Additionally, our study highlights the benefits of pre-trained models in the EEG domain, traditionally hampered by data scarcity and complexity. The demonstrated superior performance of the pretrained EEGViT model suggests that the prior knowledge learned from vast image datasets can effectively help when handling EEG data complexity. Future work may include exploring pretraining transformers for EEG data using other modalities of data. By providing an open-source version of our code\footnote{Code is available at: \url{https://github.com/ruiqiRichard/EEGEyeNet-vit}.}, we encourage further exploration and enhancements to our approach.  

Nevertheless, we recognize the limitations of our study, particularly the use of a relatively small Vision Transformer due to resource limitations. Future research with increased resources could potentially scale up the model by a factor of 10 or more, potentially leading to further performance gains. It is critical to balance model complexity, computational resources, and performance gains as we progress in this research direction.

\section{Conclusion}
In conclusion, the implementation of a Hybrid Vision Transformer (ViT) pretrained on ImageNet in an EEG regression task demonstrates the potential of Transformer-based transfer learning in neuroscience. The ViT, initially created for visual tasks, notably outperformed state-of-the-art convolution-based models in EEG data analysis.

These findings challenge the prevalent use of convolution in EEG neural networks and imply that cross-domain pretrained Transformer models may serve as valuable priors for EEG tasks with minor tweaks. This insight, especially relevant where data collection is difficult, suggests that repurposing available image data for EEG can enrich future research in computer science and neuroscience.

\newpage

\bibliographystyle{ACM-Reference-Format}
\bibliography{KDD}

\newpage
\appendix
\section{Appendix}

\subsection{Baseline Machine Learning Models}
\textbf{KNN}
K-Nearest Neighbors (KNN), a non-parametric method for regression, was implemented with a leaf size of 50. The selection of the leaf size aims to balance computational cost and prediction accuracy.\\\\
\textbf{RBF SVC/SVR}
SVMs with Radial Basis Function (RBF) kernels were used, tuned with a regularization parameter C of 1, tolerance of 1e-05 for stopping criteria, and a gamma of 0.01. These parameters were chosen to balance the trade-off between model complexity and overfitting.\\\\
\textbf{Linear Regression}
Linear Regression, a fundamental statistical method, was employed with its default parameters, providing a benchmark against more complex models.\\\\
\textbf{Lasso Regression}
Lasso Regression uses shrinkage to limit the complexity of the model and prevent overfitting. We adopted the default parameters for the Absolute Position task, given the absence of task-specific parameters.\\\\
\textbf{Elastic Net}
Elastic Net, a blend of Ridge Regression and Lasso Regression, was implemented with an alpha of 1, L1 ratio of 0.6, tolerance of 1e-05, and a gamma of 0.01. These parameters were chosen to optimize the balance between Ridge and Lasso penalties.\\\\
\textbf{Random Forest}
The Random Forest ensemble learning method was deployed with a maximum depth of 50 and 250 estimators to balance the trade-off between model complexity and computational feasibility.\\\\
\textbf{AdaBoost}
AdaBoost, an ensemble method that enhances the performance of base learners, was employed with a learning rate of 0.01 and 50 estimators. These parameters aim to avoid overfitting while maintaining a reasonable learning rate.\\\\
\textbf{XGBoost}
XGBoost, an efficient gradient boosting library, was applied with a learning rate (eta) of 0.1, 250 estimators, and a maximum depth of 10. These settings were chosen to balance training speed and model performance.\\\\
Hyperparameters can be found in \cite{kastrati2021EEGEyeNet} and \href{https://github.com/ardkastrati/EEGEyeNet}{EEGEyeNet GitHub repository}.

\subsection{Baseline Deep Learning Model Hyperparameters}

We reused the configurations and hyperparameters suggested by EEGEyeNet benchmark (see Table \ref{fig:dl_hyper}).

\begin{table}[b]
  \centering
  \includegraphics[width=\columnwidth]{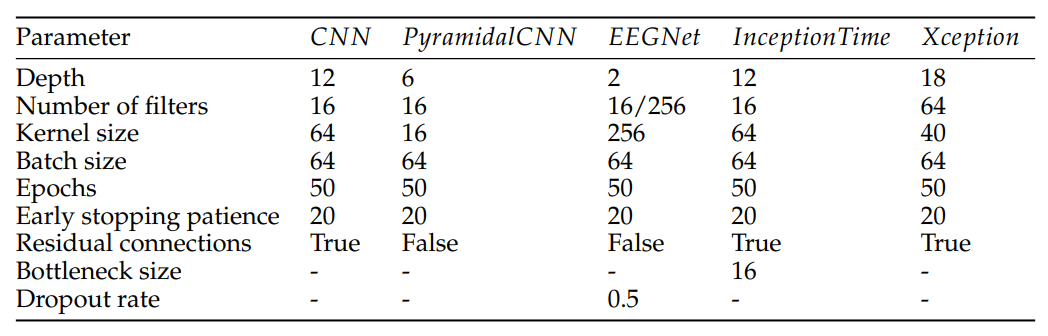}
  \caption{Hyperparameters of baseline deep learning models, suggested by EEGEyeNet \cite{kastrati2021EEGEyeNet}.}
  \label{fig:dl_hyper}
\end{table}

\subsection{Guide for reproducibility}
We make our code publicly available at our \href{https://github.com/ruiqiRichard/EEGEyeNet-vit}{GitHub repository} to facilitate further research. Details on the EEGEyeNet dataset and benchmark can be found in \cite{kastrati2021EEGEyeNet} and \href{https://github.com/ardkastrati/EEGEyeNet}{their GitHub repository}. All training data used for experimentation is publicly available on \href{https://osf.io/ktv7m/}{OSF.io}. Further information on the implementation and reproducibility of our work can be found in our GitHub repository.

\end{document}